\pdfoutput=1

\documentclass[11pt]{article}

\usepackage[preprint]{acl}

\usepackage{times}
\usepackage{latexsym}

\usepackage[T1]{fontenc}

\usepackage[utf8]{inputenc}

\usepackage{microtype}

\usepackage{inconsolata}

\usepackage{booktabs}
\usepackage{comment}
\usepackage{multirow}
\usepackage{graphicx}
\usepackage{pgfplots}
\usepackage{xurl}
\usepackage{colortbl}
\usepackage{fontawesome5}
\usepackage{hyperref}
\pgfplotsset{compat=1.18}

\definecolor{greenimg}{RGB}{16,128,3}
\definecolor{redimg}{RGB}{128,0,3}

\title{Multilingual \emph{vs} Crosslingual Retrieval of Fact-Checked Claims:\\A Tale of Two Approaches}

\author{Alan Ramponi,$^1$\thanks{These authors contributed equally to this work.} Marco Rovera,$^{1*}$ Robert Moro,$^2$ Sara Tonelli$^1$\\ 
\texttt{\{alramponi,m.rovera,satonelli\}@fbk.eu}, \texttt{robert.moro@kinit.sk} \\ \\
\textsuperscript{1} Fondazione Bruno Kessler, Trento, Italy \\
\textsuperscript{2} Kempelen Institute of Intelligent Technologies, Bratislava, Slovakia}

\begin{document}
\maketitle
\begin{abstract}
Retrieval of previously fact-checked claims is a well-established task, whose automation can assist professional fact-checkers in the initial steps of information verification. Previous works have mostly tackled the task monolingually, i.e., having both the input and the retrieved claims in the same language. However, especially for languages with a limited availability of fact-checks and in case of global narratives, such as pandemics, wars, or international politics, it is crucial to be able to retrieve claims across languages. In this work, we examine strategies to improve the multilingual \textit{and} crosslingual performance, namely selection of negative examples (in the supervised) and re-ranking (in the unsupervised setting). We evaluate all approaches on a dataset containing posts and claims in 47 languages (283 language combinations). We observe that the best results are obtained by using LLM-based re-ranking, followed by fine-tuning with negative examples sampled using a sentence similarity-based strategy. Most importantly, we show that crosslinguality is a setup with its own unique characteristics compared to the multilingual setup.\footnote{\faGithub~Code and data are publicly available at: \url{https://github.com/kinit-sk/multiclaim-emnlp2025}}
\end{abstract}

\section{Introduction}
\label{sec:intro}

Fighting online mis/disinformation is a challenging task for professional fact-checkers and human moderators alike, given that false information spreads six times faster than true information \cite{doi:10.1126/science.aap9559}. In the fact-checking pipeline, one of the tasks that can remarkably speed up operators' activity and support their work is \textit{previously fact-checked claim retrieval} (PFCR), defined as follows~\cite{pikuliak-etal-2023-multilingual,shaar-etal-2020-known}: ``Given a text making an input claim (e.g., a social media post) and a set of previously fact-checked claims, the task is to rank the fact-checked claims so that those that are the most relevant with respect to the input claim are ranked as high as possible.''

\begin{figure}[t]
    \centering
\includegraphics[width=0.99\linewidth]{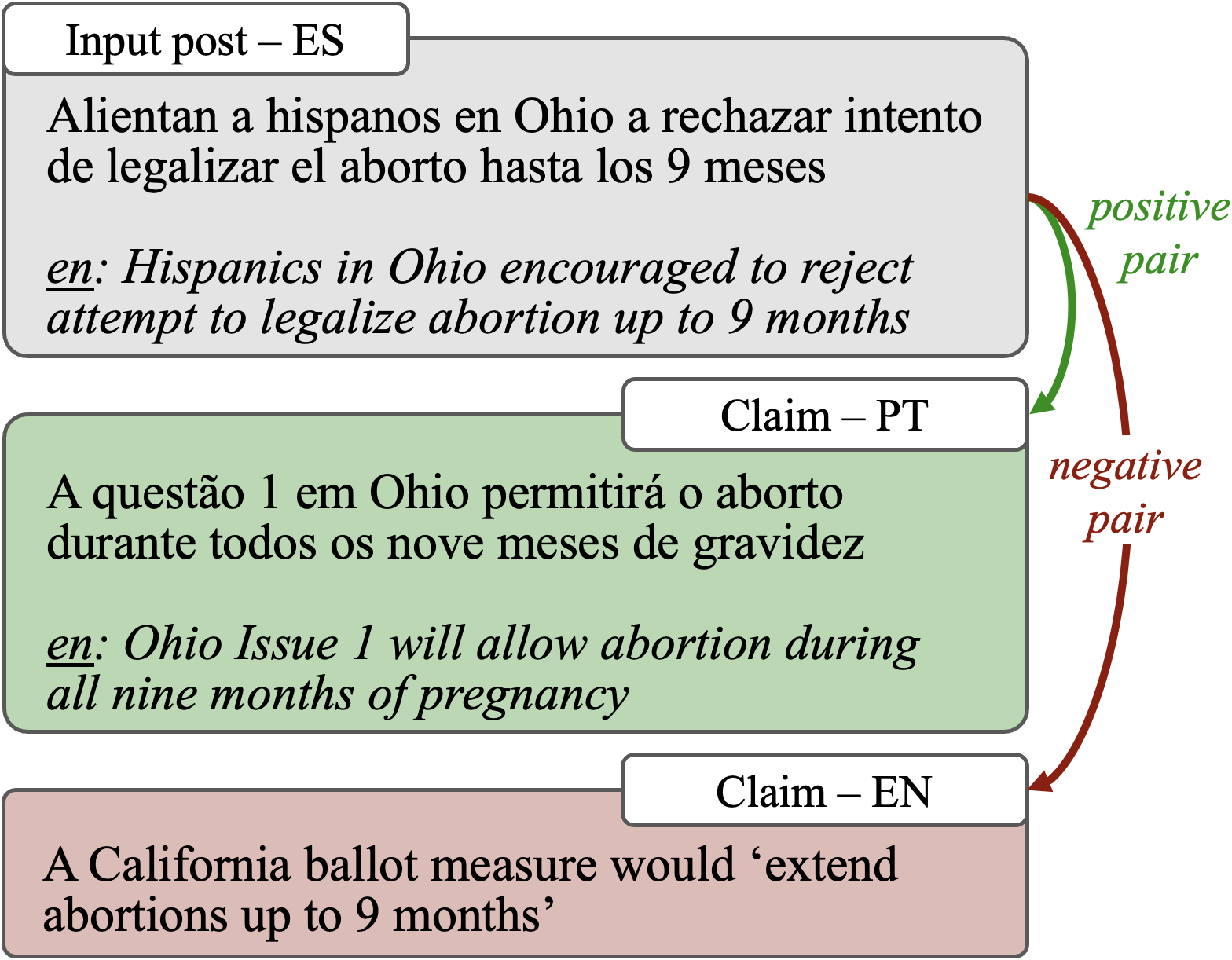}
\caption{An example of an input post paired with a previously fact-checked claim related to the same event (\textcolor{greenimg}{\textit{positive pair}}) and a claim that is similar to the input post but unrelated (\textcolor{redimg}{\textit{negative pair}}), which can be used to fine-tune a retriever. Redacted examples from the MultiClaim dataset~\cite{pikuliak-etal-2023-multilingual}.}\label{fig:examples}
\end{figure}

The task, whose goal is to avoid fact-checking the same claim again and taking advantage of existing verified knowledge, is meant to address different operators' needs. For example, a fact-checking agency may want to reuse the internal knowledge collected over the years mainly in a \textit{monolingual} fashion, i.e., when input and retrieved claims are in the same language. Another option may concern cases in which a fact-checking agency needs to verify a global narrative (e.g., related to COVID-19) and is interested in retrieving already fact-checked claims in any language, regardless of the language of the input claim. This case requires a \textit{multilingual} approach. Finally, fact-checkers may be aware of the fact that a narrative arising in their country was already present and debunked in other countries (or in other languages of their country), so they need to retrieve fact-checked claims in a language that is different from their input claim. In this case the required approach is \textit{crosslingual}. Figure~\ref{fig:examples} illustrates this last case, with an input post in Spanish, the associated fact-checked claim in Portuguese and a similar but unrelated claim in English, which can make it difficult to retrieve the correct one.

Most prior works such as \citet{shaar-etal-2020-known} and \citet{hardalov-etal-2022-crowdchecked} tackled the task monolingually focusing on English. Few recent works have proposed to extend the task to multilingual setups \cite{pikuliak-etal-2023-multilingual,panchendrarajan2025multiclaimnet}, following the initiatives by international fact-checking agencies aimed at sharing and connecting the different databases of verified information.\footnote{\url{https://www.poynter.org/ifcn/}} Only two works so far addressed the task in crosslingual setups, with the input claim being always in a different language than the retrieved fact-checked claims~\cite{pikuliak-etal-2023-multilingual,vykopal2025llms}. Both confirm that  crosslingual PFCR is a very challenging task, and show the potential of translating posts and fact-checks into English, especially for low-resource languages. 

Despite the promising performance obtained through translation, however, a range of new multilingual embedding models (e.g., mE5;~\citealp{wang2024multilingual}) and large language models (e.g., Qwen;~\citealp{yang2024qwen2technicalreport}) has been released, which could greatly benefit crosslingual tasks. Furthermore, prior works have shown a positive impact of fine-tuning these models on PFCR data, specifically when using (multiple) negative examples~\cite{pikuliak-etal-2023-multilingual, neumann-etal-2023-deep}. However, specific strategies tailored to the task and focused on crosslingual retrieval were left unexplored.

To address this gap, we compare supervised and unsupervised PFCR in multilingual and crosslingual setups, identifying the best strategies and experimental settings for the task without resorting to translation. In particular, we investigate three main aspects: 
\textbf{(RQ1)} how the different text embedding models for retrieval and re-ranking  perform on the task,  
\textbf{(RQ2)} what strategies should be used to select negative examples in a supervised framework and how these compare to unsupervised approaches, and 
\textbf{(RQ3)} what the specifics of crosslingual setup are compared to the multilingual one.
We carry out our experiments on a dataset derived from the publicly available MultiClaim dataset~\cite{pikuliak-etal-2023-multilingual}. Even if we use a pre-existing dataset, we curate it specifically for multilingual and crosslingual PFCR obtaining 283 post--fact-check language combinations. In particular, we ensure proper data splits of not only posts, but also fact-checks to prevent data contamination, which was not done in prior works and thus can be considered an additional contribution of this work. We publish our subset and data splits to ensure reproducibility.

\section{Related Work}
\label{sec:relworks}

Previously fact-checked claim retrieval, also known as claim matching or claim detection, is a standard task in the fact-checking pipeline, both manual and automated~\cite{panchendrarajan-zubiaga-2024, vykopal-etal-2024-survey}, and it can also serve as a signal of content credibility~\cite{srba-etal-2024-credsignals}. It was addressed by a series of CheckThat!~Lab shared tasks organized at CLEF in 2020~\cite{DBLP:conf/clef/Barron-CedenoEN20}, 2021~\cite{shaar_overview_2021}, and 2022~\cite{nakov_overview_2022}; most recently, it was also part of a SemEval 2025 shared task~\cite{peng-etal-2025-semeval}.

Due to the relative popularity of the task, there is a range of relevant existing datasets, as summarized in recent surveys by~\citet{panchendrarajan-zubiaga-2024} and~\citet{srba-etal-2024-credsignals}. These resources differ in the number of included languages, data volume, means of identification of pairs between input social media posts and fact-checked claims, and in sources of input posts. The highest number of input posts is in CrowdChecked ($\approx$300k; \citealp{hardalov-etal-2022-crowdchecked}) and MuMiN ($\approx$21M; \citealp{nielsen2022-mumin}); however, they either contain a high level of noise in the identified pairs (the former) or do not explicitly provide the pairs (the latter). The most linguistically diverse datasets are MuMiN with 41 languages~\cite{nielsen2022-mumin}, MultiClaim with 27 languages in posts and 39 languages in fact-checked claims~\cite{pikuliak-etal-2023-multilingual}, and MMTweets with 4 languages in posts and 11 languages in fact-checked claims~\cite{singh-etal-2024-mmtweets}. Other relevant multilingual datasets (focusing on distinct but related tasks) include the EUvsDisinfo dataset of disinformation articles matched with trustworthy articles from credible sources~\cite{leite-etal-2024-euvsdisinfo} and MultiClaimNet, a dataset that combines three existing datasets (including MultiClaim) and enriches them with identified claim clusters~\cite{panchendrarajan2025multiclaimnet}. Despite this linguistic diversity, crosslingual retrieval remains underexplored for the task, since the identified pairs of fact-checks and input posts are often in the same language.

Regarding the approaches employed for PFCR, the existing works most typically employ one or more text embedding models to encode input posts and claims for  similarity search~\cite{shaar_overview_2021, pikuliak-etal-2023-multilingual, martin2022_facter-check}. Occasionally, a re-ranker is employed~\cite{shaar_overview_2021}, the models are fine-tuned for the task~\cite{pikuliak-etal-2023-multilingual, kazemi-etal-2021-claim}, or both approaches are combined~\cite{hardalov-etal-2022-crowdchecked}. Most recently, large language models (LLMs) have been also employed in zero- and few-shot settings for PFCR using a range of prompting strategies~\cite{vykopal2025llms, pisarevskaya-zubiaga-2025-zero}, highlighting that no single strategy proved as the best overall and also that the performance is lower for low-resource languages. Finally,~\citet{neumann-etal-2023-deep} proposed to use multiple negative examples during fine-tuning of the embedding models, improving overall retrieval performance. In our work, we extend this approach by exploring and comparing a wider range of selection strategies and fine-tuned models.

\section{Dataset}
\label{sec:dataset}

To perform our experiments, we extract a subset of the MultiClaim v2 dataset.\footnote{MultiClaim v2 is an updated version of the original MultiClaim dataset~\cite{pikuliak-etal-2023-multilingual} and is available at: \url{https://doi.org/10.5281/zenodo.15413169}}
MultiClaim v2 dataset is composed of pairs of \textit{posts} and \textit{fact-checked claims}, which can be in different languages, provided that each post is linked to at least one claim (see Figure \ref{fig:examples}). It is constructed by extracting claims from fact-checking articles obtained through the Google Fact-Check Explorer API or via custom scrapers and links to posts from the \texttt{ClaimReview} schema,\footnote{\url{https://www.claimreviewproject.com/}} provided directly by the fact-checkers in the articles. Additional pairs are created through fact-checking labels on the Meta platforms (i.e., Facebook and Instagram). 

To curate a subset of data for multilingual and crosslingual PFCR, we work only with posts' text (i.e., omitting text extracted from OCRed images as it is often noisy) in their original languages. We include only languages originally represented with at least 200 posts and keep only fact-checked claims that have at least one paired post.\footnote{This leads to a cut-off of 180 posts after additional filtering to ensure non-overlapping fact-checks in the data splits.} 

Next, we split posts, fact-checked claims, and pairs into training, development, and test sets stratified by language by withholding 10\% of the data for development and 10\% for testing. We also ensure that no fact-checked claim appears in more than one split (i.e., not only posts are split, but also the search spaces differ across the splits to prevent data contamination and to have a less biased estimate of the true retrieval performance). The data distribution is shown in Table~\ref{tab:dataset_summary}. For the full list of supported languages and their distribution across posts and fact-checks, see Table~\ref{tab:data_lang_combinations} in Appendix~\ref{sec:appendix:dataset}.

With 47 languages in total (30 languages represented in the posts, 46 languages in the fact-checked claims) and 283 language combinations, the experiments reported in this paper are carried out, to the best of our knowledge, with the most linguistically diverse dataset for PFCR to date.

\begin{table}
    \centering
    \small
    \begin{tabular}{lcc}
        \toprule
         & \textbf{Multilingual} & \textbf{Crosslingual} \\
         \midrule
       \textbf{\# posts}  & \textbf{55,421} & \textbf{7,975} \\
        \emph{training set} & 44,553 & 6,343 \\
        \emph{development set} & ~~5,185 & ~~782 \\
        \emph{test set} & ~~5,683 & ~~850 \\ \midrule
       \textbf{\# fact-checks}  & \textbf{52,911}  & \textbf{7,869} \\
        \emph{training set} & 41,060 & 6,118 \\
        \emph{development set} & ~~5,706 & ~~850 \\
        \emph{test set} & ~~6,145 & ~~901 \\ \midrule
       \textbf{\# pairs}  & \textbf{63,913} & \textbf{9,066} \\
        \emph{training set} & 51,658 & 7,261 \\
        \emph{development set} & ~~5,880 & ~~876 \\
        \emph{test set} & ~~6,375 & ~~929 \\
        \bottomrule
    \end{tabular}
    \caption{Distribution of social media posts, fact-checked claims, and their pairs across train, development, and test splits for the multilingual and crosslingual setups.}
    \label{tab:dataset_summary}
\end{table}

\section{Methodology}
\label{sec:methodology}

In line with previous work, we cast PFCR as a ranking task. Specifically, given a post $p$ and a set of fact-checked claims $c_1, ..., c_n \in C$ that includes the gold claim $c_p$ for the given post, the goal is to rank the fact-checked claims so that $c_p$ ranks as high as possible. We design unsupervised and supervised approaches, and for both of them we preliminarily compute and index, for each post and fact-checked claim in the dataset, the corresponding text embedding representation.

In the \textbf{unsupervised setting}, in the first stage, we use similarity-based dense retrieval to rank the available fact checks. In a further step, we apply re-ranking to the retrieved fact-checks in order to improve accuracy. To this end, we evaluate two re-ranking techniques: cross-encoder re-ranking and LLM-based re-ranking (for a comparison, see \citealp{dejean2024thorough}). While cross-encoder re-rankers \citep{nogueira2019passage} work by comparing query-document pairs and produce a relatedness or similarity score as output, LLM-based re-rankers \citep{muennighoff2022sgpt, sun2023chatgpt} are \textit{instructed} to \textit{generate} a ranking of a set of documents given a query, thus harnessing the reasoning capabilities of the underlying model.

In the \textbf{supervised setting}, both positive and negative examples are required to fine-tune text embedding models. Although positive examples can be easily obtained from the pairs of posts and fact-checks in the dataset, there is not a single and well-established way to select negative examples for training. Previous work has mainly focused on random sampling~\citep{pikuliak-etal-2023-multilingual}, i.e., creating a negative example by pairing a given post with a fact-checked claim randomly picked from those not associated with the input post.
Albeit straightforward, such a strategy leads to rapid saturation of the training set because negative and positive examples can be easily discriminated after a few training iterations.\footnote{This is also likely to happen if we consider the set of (filtered) claims without any paired post as negative examples.} We therefore design two approaches to mitigate training set saturation during fine-tuning through the sampling of challenging negative pairs. 

We experiment by sampling topically relevant (\texttt{topic}) and semantically similar (\texttt{similarity}) negative pairs, using the \texttt{random} strategy as a baseline for comparison. An example of negative pair based on \texttt{similarity} is reported in Figure~\ref{fig:examples}, with the fact-checked claim sharing terms such as `abortion(s)' and `9 months' with the input post. This example is very challenging for the model, since it should be able to discriminate between measures on abortion proposed in California and in Ohio. On the contrary, pairs sampled randomly may be fully unrelated, and the classifier may learn to discriminate them simply because they are about different topics.
Furthermore, we investigate the impact of using a varying number $k$ of negative examples on performance across our negative sampling strategies and the random sampling baseline.

For \texttt{topic}, we compute text embeddings for both posts and fact-checked claims, cluster them through topic modeling,\footnote{We use BERTopic~\citep{grootendorst2022bertopic} for topic modeling and \texttt{multilingual-e5-large} for text embeddings.} and then create $k$ negative pairs by selecting and associating to the post a number $k$ of fact-checks at random from within the same cluster to which the post belongs.\footnote{In the case there are $\leq k$ fact-checked claims in the cluster, we draw them from a cluster of uncategorized fact-checks (i.e., without assigned topic) as a fallback. This is similar in spirit to random selection.} For \texttt{similarity}, we compute the cosine similarity between each post and all the fact-checked claims and create $k$ negative pairs by associating each post to the top-$k$ most similar fact-checks. In creating all negative pairs, we ensure that each sampled fact-checked claim is not already associated with the post as a positive pair. To avoid costly computation during fine-tuning and ensure reproducibility, we create negative examples across strategies offline and serialize them to be used at training time. Details and hyper-parameters are in Section~\ref{sec:eval:supervised}.

\section{Experimental Setup}
\label{sec:experiments}

We experiment with unsupervised retrieval with and without re-ranking (Section~\ref{sec:eval:unsupervised}) and supervised retrieval using three negative sampling strategies and a varying amount of negatives (Section~\ref{sec:eval:supervised}). We evaluate these approaches on the test set in two main data settings: \emph{i)} a \textit{multilingual} setting, without distinction between monolingual and crosslingual pairs, and \emph{ii}) a \textit{crosslingual} setting, including only post--fact-check pairs in different languages. 
We rely on two widely used text embedding models of varying size to compare the two approaches, namely \texttt{multilingual-e5-large} and \texttt{paraphrase-multilingual-mpnet-base-v2}, and further compare the performance of 14 additional models in the unsupervised setting.\footnote{We select two widely-used models for the supervised setting due to computational constraints. \texttt{paraphrase-multilingual-mpnet-base-v2} was selected as it was the best multilingual model in \citet{pikuliak-etal-2023-multilingual}.} All are evaluated using \textit{Pair Success at 10} (S@10) and \textit{Mean Reciprocal Rank at 10} (MRR@10). The former measures, for each post, whether the paired claim appears in the top-10 retrieved results, while the latter also considers its position (i.e., rank).

\subsection{Unsupervised Text Embedding Models}
\label{sec:eval:unsupervised}
We select 16 among the most recent, top performing multilingual embedding models, resulting in a diverse set of architectures, pre-training techniques, number of parameters (278M--7B) and data, including the use of synthetic data. Models have been selected based on their performance for the `Retrieval' task in the Multilingual Text Embedding Benchmark\footnote{\url{https://huggingface.co/spaces/mteb/leaderboard} (last visited: 05/13/2025).}~\cite[MTEB;][]{enevoldsen2025mmteb}. Experiments\footnote{All experiments have been conducted on a single Nvidia A40 GPU, equipped with 48 GB RAM.} are run on both the \textit{test} set (in order to ensure fair comparison with the supervised approach) and the \textit{full} set of claims in the dataset. This enables us to estimate the scalability capabilities of the models. In fact, while in the test set the set of claims \textit{C} consists of over 6,100 claims (Table~\ref{tab:dataset_summary}), in the full set the model is required to rank over 52,000 claims for each input post. To this end, we test traditional models like \texttt{paraphrase-multilingual-mpnet-base-v2} and \texttt{multilingual-MiniLM-L12-v2}~\cite{reimers-2019-sentence-bert}, encoder-based models like \texttt{bge-m3}~\cite{chen-etal-2024-m3}, \texttt{jina-embeddings-v3}~\cite{sturua2024jina}, \texttt{multilingual-e5-large}~\cite{wang2024multilingual}, \texttt{snowflake-arctic-embed-l-v2.0}~\cite{yu2024arctic}, and \texttt{gte-multilingual-base}~\cite{zhang2024mgte}, as well as LLM-based models like \texttt{bge-multilingual-gemma2}~\cite{chen-etal-2024-m3}, \texttt{E5-mistral-7b}~\cite{wang2024improving}, \texttt{Linq-Embed-Mistral}~\cite{choi2024linqembedmistral}, \texttt{
KaLM-embedding-multilingual-mini-v1}~\cite{hu2025kalm}, and \texttt{gte-Qwen2-7B-instruct}~\cite{li2023generaltextembeddingsmultistage}. Where available, we include instructed versions.  While prioritizing open source and research models, we also test OpenAI's \texttt{text-embedding-3-large}~\cite{openai_textembedding3large_2024} as a benchmark.
For re-rankers, we use \texttt{bge-reranker-v2-m3}\footnote{\url{https://huggingface.co/BAAI/bge-reranker-v2-m3}} \cite{li2023makinglargelanguagemodels} (cross-encoder) and \texttt{RankGPT}~\cite{sun2023chatgpt} (LLM-based). We test the RankGPT re-ranker on the three best performing embedding models in retrieval (on the test set).\footnote{This was motivated by computational and funding reasons, as each test set experiment costed approximately \$50. The three models have been chosen because of their high performance in pure retrieval and to enable comparison with the supervised approach (\texttt{multilingual-e5-large}).} The only hyper-parameter for re-ranking, i.e., the top-\textit{n} claims to re-rank, has been explored in a preliminary phase, experimenting with $n \in [20,30,50,100]$. We found $20$ and $30$ to be the best values. To not introduce additional burden to our already composite experimental setup, throughout the paper we report re-ranking results with top-$n=30$.

\subsection{Supervised Text Embedding Models}
\label{sec:eval:supervised}

We employ \texttt{multilingual-e5-large} and \texttt{paraphrase-multilingual-mpnet-base-v2} as our pretrained text embedding models to fine-tune. We select them since they are widely used models, represent varied parameter sizes (278M--560M), and can be fine-tuned on an average GPU\footnote{We rely on a single Tesla V100-SXM2-32GB GPU.} without the need for a costly computing infrastructure. We use multiple negatives ranking loss and perform hyper-parameter tuning for selecting the best value for the learning rate (among 1e-9, 5e-9, 1e-8, 5e-8, 1e-7), the batch size (4, 8, 16), and warm-up steps (800, 1,600) based on S@10 multilingual performance on the development set for both models using all negative sampling approaches. We select 1e-8 as the learning rate, 8 as batch size, and 1,600 as warm-up steps. The full list of hyper-parameters is reported in Appendix~\ref{sec:appendix:supervised}.

To select the model configurations to be used for test set evaluation, we further investigate which number $k$ of negative examples provides the best overall performance across models and the three negative sampling strategies. For stability, we run each model configuration three times, each with a different seed, and report average results. We experiment with $k \in [1,2,3,4,5,10]$ and find that using $k=10$ negative examples provides the best overall performance for all strategies on the development set. As shown in Figure~\ref{fig:negatives-k-and-strategies-e5} for the fine-tuned \texttt{multilingual-e5-large} model, \texttt{similarity} consistently outperforms both the \texttt{topic} and \texttt{random} strategies according to the S@10 multilingual score, with the best results obtained when using 10 negatives. These results are consistent with those obtained when fine-tuning \texttt{paraphrase-multilingual-mpnet-base-v2} (cf. Figure~\ref{fig:negatives-k-and-strategies-mpnet} in Appendix~\ref{sec:appendix:supervised}) and motivate our selection of $k=10$ for test set evaluation.

\begin{figure}[!t]
    \resizebox{\linewidth}{!}{%
    
        \begin{tikzpicture}
            \begin{axis}[
                xlabel={$k$ (\# negative examples)},
                ylabel={S@10},
                xmin=0.5, xmax=10.5,
                ymin=0.832, ymax=0.858,
                xtick={1,2,3,4,5,10},
                ytick={0.85,0.84},
                height=5cm,
                width=9.5cm,
                legend style={at={(0.5,-0.38)}, anchor=north, legend columns=-1},
                ymajorgrids=true,
                grid style=dashed,
            ]

            \addplot[
                color=red!75!white,
                mark=triangle*,
                ]
                coordinates {
                (1,0.8365)(2,0.8382)(3,0.8405)(4,0.8424)(5,0.8433)(10,0.8467)
                };
            \addlegendentry{random}

            \addplot[
                color=green!60!black,
                mark=square*,
                ]
                coordinates {
                (1,0.8451)(2,0.8485)(3,0.8488)(4,0.8505)(5,0.8516)(10,0.8535)
                };
            \addlegendentry{similarity}

            \addplot[
                color=blue!60!black,
                mark=*,
                ]
                coordinates {
                (1,0.8436)(2,0.8456)(3,0.8473)(4,0.8481)(5,0.8486)(10,0.8498)
                };
            \addlegendentry{topic}

            \draw[dashed] (1,0.8343) -- (10,0.8343);
            \node at (8,0.8360) {\emph{without fine-tuning}};
            \end{axis}
        \end{tikzpicture}
    
}%
	\caption{\textbf{Multilingual} S@10 performance across negative sampling strategies and number of negative examples $k$ for the fine-tuned \texttt{multilingual-e5-large} model on the \textbf{development} set. Reported results are averaged over three runs using different seeds. The dashed line indicates results when no fine-tuning is conducted.
    }
	\label{fig:negatives-k-and-strategies-e5}
\end{figure}

\section{Results and Discussion}

In this section, we present the results on the test set of unsupervised (Section~\ref{sec:results:unsupervised}) and supervised (Section~\ref{sec:results:supervised}) approaches, along with a detailed discussion. We then compare the approaches and discuss limitations and future directions (Section~\ref{sec:results:comparison}).

\subsection{Unsupervised Results}
\label{sec:results:unsupervised}

\begin{figure*}[!ht]
\centering
\begin{tikzpicture}
\begin{axis}[
    ybar,
    bar width=6pt,
    width=16cm,
    height=5cm,
    at={(0,7.6cm)}, %
    enlargelimits=0.06,
    ylabel={MULTI},
    ymin=0.5,
    ymax=0.95,
    symbolic x coords={Linq-Embed-Mistral, multilingual-MiniLM-L12-v2, text-embedding-3-small, e5-mistral-7b-instruct, multilingual-e5-large-instruct, KaLM-multilingual-mini-instruct-v1, paraphrase-multilingual-mpnet-base-v2, SFR-Embedding-Mistral, multilingual-e5-large, gte-Qwen2-7B-instruct, bge-m3, jina-embeddings-v3, snowflake-arctic-embed-l-v2.0, text-embedding-3-large, gte-multilingual-base, bge-multilingual-gemma2},
    xtick=\empty, %
    nodes near coords,
    nodes near coords style={/pgf/number format/fixed, /pgf/number format/precision=4, font=\scriptsize, rotate=90, anchor=west, color=black, yshift=0.5pt, xshift=-2pt},
    legend style={at={(0.5,1.2)}, anchor=north, legend columns=-1, font=\footnotesize},
    ylabel style={font=\small},
    xlabel={}, %
    xlabel style={font=\small},
    tick label style={font=\small},
    major tick length=0.2cm,
    ]
\addplot+[style={fill=violet!110!black}]
    coordinates {(Linq-Embed-Mistral, 0.7199) (multilingual-MiniLM-L12-v2, 0.5062) (text-embedding-3-small, 0.7229) (e5-mistral-7b-instruct, 0.6798) (multilingual-e5-large-instruct, 0.7311) (KaLM-multilingual-mini-instruct-v1, 0.7155) (paraphrase-multilingual-mpnet-base-v2, 0.5690) (SFR-Embedding-Mistral, 0.7309) (multilingual-e5-large, 0.7475) (gte-Qwen2-7B-instruct, 0.7155) (bge-m3, 0.6952) (jina-embeddings-v3, 0.7137) (snowflake-arctic-embed-l-v2.0, 0.7344) (text-embedding-3-large, 0.7909) (gte-multilingual-base, 0.7228) (bge-multilingual-gemma2, 0.7613)};
\addplot+[style={fill=orange!110!black}]
    coordinates {(Linq-Embed-Mistral, 0.7174) (multilingual-MiniLM-L12-v2, 0.6038) (text-embedding-3-small, 0.7035) (e5-mistral-7b-instruct, 0.7223) (multilingual-e5-large-instruct, 0.7341) (KaLM-multilingual-mini-instruct-v1, 0.7275) (paraphrase-multilingual-mpnet-base-v2, 0.6554) (SFR-Embedding-Mistral, 0.7296) (multilingual-e5-large, 0.7323) (gte-Qwen2-7B-instruct, 0.7219) (bge-m3, 0.7031) (jina-embeddings-v3, 0.7120) (snowflake-arctic-embed-l-v2.0, 0.7130) (text-embedding-3-large, 0.7321) (gte-multilingual-base, 0.7161) (bge-multilingual-gemma2, 0.7298) };
\addplot+[style={fill=cyan!130!black}]
    coordinates {(Linq-Embed-Mistral,) (multilingual-MiniLM-L12-v2,) (text-embedding-3-small,) (e5-mistral-7b-instruct,) (multilingual-e5-large-instruct,) (KaLM-multilingual-mini-instruct-v1,) (paraphrase-multilingual-mpnet-base-v2,) (SFR-Embedding-Mistral,) (multilingual-e5-large, 0.8042) (gte-Qwen2-7B-instruct,) (bge-m3,) (jina-embeddings-v3,) (snowflake-arctic-embed-l-v2.0,) (text-embedding-3-large, 0.8212) (gte-multilingual-base,) (bge-multilingual-gemma2, 0.8104)};
\legend{Retrieval, Cross-encoder re-ranking, LLM-based re-ranking}
\node[anchor=north west, font=\bfseries] at (rel axis cs:0,1) {(a)};
\end{axis}

\begin{axis}[
    ybar,
    bar width=6pt,
    width=16cm,
    height=5cm,%
    at={(0,3.8cm)},
    enlargelimits=0.06,
    ylabel={CROSS},
    ymin=0.1,
    ymax=0.82,%
    symbolic x coords={Linq-Embed-Mistral, multilingual-MiniLM-L12-v2, text-embedding-3-small, e5-mistral-7b-instruct, multilingual-e5-large-instruct, KaLM-multilingual-mini-instruct-v1, paraphrase-multilingual-mpnet-base-v2, SFR-Embedding-Mistral, multilingual-e5-large, gte-Qwen2-7B-instruct, bge-m3, jina-embeddings-v3, snowflake-arctic-embed-l-v2.0, text-embedding-3-large, gte-multilingual-base, bge-multilingual-gemma2},
    xtick=\empty,
    x tick label style={rotate=40, anchor=east, font=\scriptsize},
    nodes near coords,
    nodes near coords style={/pgf/number format/fixed, /pgf/number format/precision=4, font=\scriptsize, rotate=90, anchor=west, color=black, yshift=0.5pt, xshift=-2pt},
    legend style={at={(0.5,1.2)}, anchor=north, legend columns=-1, font=\footnotesize},
    xlabel={},
    ylabel style={font=\small},
    xlabel style={font=\large},
    tick label style={font=\small},
    major tick length=0.2cm,
    ]
\addplot+[style={fill=violet!110!black}]
    coordinates {(Linq-Embed-Mistral, 0.2147) (multilingual-MiniLM-L12-v2, 0.2479) (text-embedding-3-small, 0.2488) (e5-mistral-7b-instruct, 0.2837) (multilingual-e5-large-instruct, 0.3318) (KaLM-multilingual-mini-instruct-v1, 0.3365) (paraphrase-multilingual-mpnet-base-v2, 0.3547) (SFR-Embedding-Mistral, 0.3744) (multilingual-e5-large, 0.4323) (gte-Qwen2-7B-instruct, 0.4385) (bge-m3, 0.4844) (jina-embeddings-v3, 0.5048) (snowflake-arctic-embed-l-v2.0, 0.5108) (text-embedding-3-large, 0.5119) (gte-multilingual-base, 0.5443) (bge-multilingual-gemma2, 0.5446)};
\addplot+[style={fill=orange!110!black}]
    coordinates {(Linq-Embed-Mistral, 0.341) (multilingual-MiniLM-L12-v2, 0.3512) (text-embedding-3-small, 0.3506) (e5-mistral-7b-instruct, 0.4501) (multilingual-e5-large-instruct, 0.4998) (KaLM-multilingual-mini-instruct-v1, 0.49) (paraphrase-multilingual-mpnet-base-v2, 0.4482) (SFR-Embedding-Mistral, 0.4968) (multilingual-e5-large, 0.5442) (gte-Qwen2-7B-instruct, 0.5283) (bge-m3, 0.5195) (jina-embeddings-v3, 0.5433) (snowflake-arctic-embed-l-v2.0, 0.5489) (text-embedding-3-large, 0.5418) (gte-multilingual-base, 0.5592) (bge-multilingual-gemma2, 0.5608)};
\addplot+[style={fill=cyan!130!black}]
    coordinates {(Linq-Embed-Mistral,) (multilingual-MiniLM-L12-v2,) (text-embedding-3-small,) (e5-mistral-7b-instruct,) (multilingual-e5-large-instruct,) (KaLM-multilingual-mini-instruct-v1,) (paraphrase-multilingual-mpnet-base-v2,) (SFR-Embedding-Mistral,) (multilingual-e5-large, 0.614) (gte-Qwen2-7B-instruct,) (bge-m3,) (jina-embeddings-v3,) (snowflake-arctic-embed-l-v2.0,) (text-embedding-3-large, 0.6328) (gte-multilingual-base,) (bge-multilingual-gemma2, 0.6559)};
    \node[anchor=north west, font=\bfseries] at (rel axis cs:0,1) {(b)};
\end{axis}

\begin{axis}[
    ybar,
    bar width=6pt,
    width=16cm,
    height=5cm,
    enlargelimits=0.06,
    ylabel={CROSS NO ENG},
    ymin=0.2,
    ymax=0.75,
    symbolic x coords={Linq-Embed-Mistral, multilingual-MiniLM-L12-v2, text-embedding-3-small, e5-mistral-7b-instruct, multilingual-e5-large-instruct, KaLM-multilingual-mini-instruct-v1, paraphrase-multilingual-mpnet-base-v2, SFR-Embedding-Mistral, multilingual-e5-large, gte-Qwen2-7B-instruct, bge-m3, jina-embeddings-v3, snowflake-arctic-embed-l-v2.0, text-embedding-3-large, gte-multilingual-base, bge-multilingual-gemma2},
    xtick=data, %
    nodes near coords,
    nodes near coords style={/pgf/number format/fixed, /pgf/number format/precision=4, font=\scriptsize, rotate=90, anchor=west, color=black, yshift=0.5pt, xshift=-2pt},
    legend style={at={(0.5,1.2)}, anchor=north, legend columns=-1, font=\footnotesize},
    ylabel style={font=\small},
    xlabel={}, %
    x tick label style={rotate=45, anchor=east, font=\scriptsize},
    tick label style={font=\small},
    major tick length=0.2cm,
    ]
\addplot+[style={fill=violet!110!black}]
    coordinates {(Linq-Embed-Mistral, 0.3668) (multilingual-MiniLM-L12-v2, 0.279) (text-embedding-3-small, 0.3726) (e5-mistral-7b-instruct, 0.3562) (multilingual-e5-large-instruct, 0.4018) (KaLM-multilingual-mini-instruct-v1, 0.374) (paraphrase-multilingual-mpnet-base-v2, 0.3975) (SFR-Embedding-Mistral, 0.4576) (multilingual-e5-large, 0.4543) (gte-Qwen2-7B-instruct, 0.5057) (bge-m3, 0.5032) (jina-embeddings-v3, 0.4603) (snowflake-arctic-embed-l-v2.0, 0.5408) (text-embedding-3-large, 0.4993) (gte-multilingual-base, 0.4972) (bge-multilingual-gemma2, 0.5291)};
\addplot+[style={fill=orange!110!black}]
    coordinates {(Linq-Embed-Mistral, 0.47) (multilingual-MiniLM-L12-v2, 0.3895) (text-embedding-3-small, 0.5037) (e5-mistral-7b-instruct, 0.4675) (multilingual-e5-large-instruct, 0.4916) (KaLM-multilingual-mini-instruct-v1, 0.512) (paraphrase-multilingual-mpnet-base-v2, 0.508) (SFR-Embedding-Mistral, 0.4907) (multilingual-e5-large, 0.5653) (gte-Qwen2-7B-instruct, 0.5408) (bge-m3, 0.5302) (jina-embeddings-v3, 0.5652) (snowflake-arctic-embed-l-v2.0, 0.5797) (text-embedding-3-large, 0.5629) (gte-multilingual-base, 0.5556) (bge-multilingual-gemma2, 0.5562) };
\addplot+[style={fill=cyan!130!black}]
    coordinates {(Linq-Embed-Mistral,) (multilingual-MiniLM-L12-v2,) (text-embedding-3-small,) (e5-mistral-7b-instruct,) (multilingual-e5-large-instruct,) (KaLM-multilingual-mini-instruct-v1,) (paraphrase-multilingual-mpnet-base-v2,) (SFR-Embedding-Mistral,) (multilingual-e5-large, 0.5797) (gte-Qwen2-7B-instruct,) (bge-m3,) (jina-embeddings-v3,) (snowflake-arctic-embed-l-v2.0,) (text-embedding-3-large, 0.5404) (gte-multilingual-base,) (bge-multilingual-gemma2, 0.5706)};
    \node[anchor=north west, font=\bfseries] at (rel axis cs:0,1) {(c)};
\end{axis}

\end{tikzpicture}
\caption{\textbf{Test} set performance (MRR@10) for retrieval and re-ranking across models in \textbf{multilingual} (a), \textbf{crosslingual} (b) and \textbf{crosslingual no eng} (i.e., without English) (c) settings. Results refer to 6,375 pairs (a), 929 pairs (b), and 118 pairs (c), and are sorted by crosslingual \textit{retrieval} performance.}
\label{fig:mrr_retrieve_rerank_performance_test}
\end{figure*}

\paragraph{Retrieval}
\label{sec:eval:unsupervised:retrieval}
In the \textit{multilingual} setting (Figure \ref{fig:mrr_retrieve_rerank_performance_test} (a), purple bars) \texttt{text-embedding-3-large} leads the ranking, with \texttt{bge-multilingual-gemma2} as the best open source model and \texttt{multilingual-e5-large} as the best lightweight model. This trend is consistent across the test and the full data settings, as well as across both metrics (for all details on multilingual evaluation, see Table  \ref{tab:model_comparison_multilingual} and Figure \ref{fig:s10_retrieve_rerank_performance_test_multi} in Appendix \ref{sec:appendix:unsupervised}). 
The \textit{crosslingual} setting (Figure \ref{fig:mrr_retrieve_rerank_performance_test} (b), purple bars) which proved more challenging, provides a different picture, with \texttt{bge-multilingual-gemma2} scoring the best result on the test set and \texttt{gte-multilingual-base} ranking first on the full set of claims. The two models outperform  \texttt{text-embedding-3-large} by a margin of 3.27 and 3.58 MRR@10 points respectively. Also in this case, results are consistent across the two adopted metrics (for all details on crosslingual evaluation, see Table \ref{tab:model_comparison_crosslingual} and  Figure \ref{fig:s10_retrieve_rerank_performance_test_cross} in Appendix \ref{sec:appendix:unsupervised}). In order to better understand models' performance, we further filtered the results by removing pairs containing a post or a claim in English, which is the dominant language in the dataset. In this setting, open-source models like \texttt{snowflake-arctic-embed-l-v2.0} and \texttt{bge-multilingual-gemma2} significantly outperform \texttt{text-embedding-3-large}. Results are reported in Figure \ref{fig:mrr_retrieve_rerank_performance_test} (c), purple bars.

\paragraph{Re-ranking}
\label{sec:eval:unsupervised:re-ranking}
For cross-encoder re-ranking, results show different effects across models and linguistic settings. In the \textit{multilingual} setting (see Figure \ref{fig:mrr_retrieve_rerank_performance_test} (a), orange bars), the effect of re-ranking is nuanced: in terms of MRR@10, and compared to pure retrieval, cross-encoder re-ranking does not yield \textit{better} rankings, except for low-performing models (for the complete results on multilingual re-ranking, please refer to Table \ref{tab:model_comparison_multilingual} and Figures \ref{fig:s10_rerank_performance_test_multi}, \ref{fig:mrr_rerank_performance_test_multi},
\ref{fig:s10_rerank_performance_full_multi}  and \ref{fig:mrr_rerank_performance_full_multi} in Appendix \ref{sec:appendix:unsupervised}). This trend is also reflected by the S@10 metric, which shows that re-ranking boosts the performance of weaker and moderately increases performance in mid-performing models, but it reduces the performance of high-performing models. In fact, in the multilingual setup, cross-encoder re-ranking has an average MRR@10 gain of 0.60 and 0.02 points on the test and the full set respectively, while S@10 shows an average increase of 1.37 and 1.77 points, respectively.
The effects of re-ranking, by contrast, emerge much more clearly in the \textit{crosslingual} setting (Figure \ref{fig:mrr_retrieve_rerank_performance_test} (b), orange and cyan bars). In fact, in this case cross-encoder re-ranking proves effective in boosting performance across all models, both in terms of MRR@10 and S@10, thus yielding better quality rankings. Indeed, the  average gain is 8.11 and 7.04 points for MRR@10 (see details in Tables \ref{fig:mrr_rerank_performance_test_cross} and \ref{fig:mrr_rerank_performance_full_cross} in Appendix \ref{sec:appendix:unsupervised}) and 5.32 and 4.93 points for S@10 (Tables \ref{fig:s10_rerank_performance_test_cross} and \ref{fig:s10_rerank_performance_full_cross} in Appendix \ref{sec:appendix:unsupervised}). A comparable performance increase can be observed in the crosslingual setup when English data are excluded (see Figure \ref{fig:mrr_retrieve_rerank_performance_test} (c), orange bars), with gains of 8.08 (test) and 5.91 (full) MRR@10 points. LLM-based re-ranking, on the other hand, demonstrates superior performance in multilingual, crosslingual, and crosslingual without English setups. It results in an average increase of 4.54, 13.80, and 6.93 MRR@10 points respectively, besides yielding the best overall results (Figure \ref{fig:mrr_retrieve_rerank_performance_test}, cyan bars).

To better understand the results, we also conducted a correlation analysis between \textit{a)} the embedding dimension and \textit{b)} the number of parameters of each embedding model on the one hand, and the observed performance on the other. The analysis, conducted using Pearson's \textit{r} coefficient, shows that there is no correlation between these variables and the observed performance (Table \ref{tab:correlation_analysis} in Appendix \ref{sec:appendix:unsupervised}).

Overall, our unsupervised experiments highlight the following relevant trends: \textit{a)} with the exception of \texttt{bge-multilingual-gemma2}, which shows high performance in both contexts, models that perform best in the multilingual setting do not always perform equally well in the crosslingual setting, and vice versa. This, once again, reflects the uniqueness of the crosslingual context. Moreover, \textit{b)} smaller, encoder-only embedding models (like \texttt{multilingual-e5-large} and \texttt{gte-multilingual-base}) often challenge or even outperform larger decoder-only models (like \texttt{bge-multilingual-gemma2} or \texttt{text-embedding-3-large}) in the task of claim retrieval, in particular in the crosslingual setup. When combined with the above-mentioned correlation analysis, this indicates that model size, embedding dimensionality, and model architecture do not impact significantly performance, suggesting that other factors, such as language coverage, data variety, and the used pre-training method may have greater influence on results. %
Finally, \textit{c)} re-ranking proves effective in many cases, although its contribution is more evident in the crosslingual setup. Albeit tested on a limited number of models, LLM-based re-ranking yields major performance improvements with respect to cross-encoder re-ranking, but at a higher computational cost.

\subsection{Supervised Results}
\label{sec:results:supervised}

For supervised experiments, we observe that fine-tuning the models using \texttt{similarity} as a negative sampling strategy consistently improves the MRR@10 performance over the \texttt{topic} approach as well as the \texttt{random} selection baseline in both \emph{multilingual} and \emph{crosslingual} settings (Table~\ref{tab:supervised_mrr}). This is further confirmed by S@10 scores (Table~\ref{tab:supervised_success_at_10} in Appendix~\ref{sec:appendix:additional}). Among the two models, \texttt{multilingual-e5-large} provides the best overall results across all strategies, showing an improvement of 1.45 and 5.31 MRR@10 points and 0.94 and 4.96 S@10 points over \texttt{random} when using the \texttt{similarity} strategy in multilingual and crosslingual setups, respectively. Although also the \texttt{topic} strategy outperforms the commonly employed \texttt{random} selection baseline~\citep{pikuliak-etal-2023-multilingual}, it still lags behind the \texttt{similarity} approach (by 0.41 and 0.54 MRR@10 points and 0.25 and 0.38 S@10 points in multilingual and crosslingual setups). The \texttt{similarity} sampling strategy therefore appears to be a viable approach for selecting hard negative examples for fine-tuning due to stable performance improvements over the other methods. Furthermore, it does not rely on the costly computation of topic clusters of the \texttt{topic} strategy to draw negative examples from (Section~\ref{sec:methodology}).

Looking at the crosslingual MRR@10 performance across strategies and the number of negative examples (see Figure~\ref{fig:negatives-n-and-strategies-e5-test-mrr}), we observe a large performance gain between the proposed strategies and the \texttt{random} baseline and the unsupervised setting over all $k$ values (results using S@10 show the same trend, see Figure~\ref{fig:negatives-n-and-strategies-e5-test} in Appendix~\ref{sec:appendix:supervised}). The \texttt{topic} strategy seems more effective when few negative examples are used for fine-tuning (i.e., $\leq5$), whereas \texttt{similarity} confirms to be the most effective approach in the PFCR task when sampling more negative examples for each pair. Overall, compared to using \texttt{multilingual-e5-large} in an unsupervised fashion, fine-tuning it with just 10 negative examples selected using the \texttt{similarity} strategy leads to a 
crosslingual MRR@10 score of 0.4947 (+6.24 points) and a crosslingual S@10 score of 0.7076 (+7.52 points). This indicates that our approach is effective even in the more challenging crosslingual setup. When excluding the highly-represented English language in the data from crosslingual evaluation (i.e., ``crosslingual (no eng)'' in Table~\ref{tab:supervised_mrr}), we observe the same trend, with \texttt{topic} and \texttt{similarity} achieving large performance gains compared to \texttt{random}.

\begin{table*}[ht]
\centering
\small
    \begin{tabular}{lllll}
        \toprule
        \textbf{Model} & \textbf{Strategy} & \textbf{Multilingual} & \textbf{Crosslingual} & \textbf{Crosslingual (no eng)} \\
        \midrule
        \multirow[c]{6}{*}{\shortstack[l]{\texttt{multilingual-}\\\texttt{e5-large}}}
        & \texttt{\textcolor{orange!100!black}{re-rank}} & 0.7323 & 0.5442 & 0.5653 \\
        & \texttt{\textcolor{violet!100!black}{retrieve}} & 0.7475 & 0.4323 & 0.4543 \\
        & \texttt{\textcolor{red!60!black}{random}} & 0.7623$_{\pm 0.0002}$ & 0.4416$_{\pm 0.0007}$ & 0.4810$_{\pm 0.0017}$ \\
        & \texttt{\textcolor{blue!60!black}{topic}} & 0.7727$_{\pm 0.0001}$ & 0.4893$_{\pm 0.0004}$ & 0.5349$_{\pm 0.0000}$ \\
        & \texttt{\textcolor{green!60!black}{similarity}} & 0.7768$_{\pm 0.0001}$ & 0.4947$_{\pm 0.0001}$ & 0.5269$_{\pm 0.0004}$ \\
        & \texttt{\textcolor{cyan!100!black}{llm re-rank}} & \textbf{0.8042} & \textbf{0.6140} & \textbf{0.5797} \\
        \midrule
        \multirow[c]{5}{*}{\shortstack[l]{\texttt{paraphrase-}\\\texttt{multilingual-}\\\texttt{mpnet-base-v2}}} 
        & \texttt{\textcolor{red!60!black}{random}} & 0.5202$_{\pm 0.0003}$ & 0.3177$_{\pm 0.0001}$ & 0.3518$_{\pm 0.0000}$ \\
        & \texttt{\textcolor{blue!60!black}{topic}} & 0.5412$_{\pm 0.0002}$ & 0.3367$_{\pm 0.0001}$ & 0.3729$_{\pm 0.0004}$ \\
        & \texttt{\textcolor{green!60!black}{similarity}} & 0.5537$_{\pm 0.0002}$ & 0.3500$_{\pm 0.0003}$ & 0.3961$_{\pm 0.0002}$ \\
        & \texttt{\textcolor{violet!100!black}{retrieve}} & 0.5690 & 0.3547 & 0.3975 \\
        & \texttt{\textcolor{orange!100!black}{re-rank}} & \textbf{0.6485} & \textbf{0.4482} & \textbf{0.5080} \\
        \bottomrule
    \end{tabular}
    \caption{\textbf{Multilingual}, \textbf{crosslingual}, and \textbf{crosslingual (no eng)} MRR@10 performance across negative sampling strategies (\texttt{random}, \texttt{topic}, \texttt{similarity}) for fine-tuned models on the \textbf{test} set compared to unsupervised strategies (\texttt{retrieve}, cross-encoder \texttt{re-rank}, \texttt{llm re-rank}). Results are ordered by increasing multilingual score; for the supervised setup, we report averages with standard deviation over three runs using different seeds.}
    \label{tab:supervised_mrr}
\end{table*}

\begin{table}[ht]
\centering
\small
    \resizebox{1\columnwidth}{!}{
    \begin{tabular}{llll}
        \toprule
        \textbf{Language} & \textbf{Strategy} & \textbf{Monolingual} & \textbf{Crosslingual} \\
        \midrule
        \multirow[c]{6}{*}{\shortstack[l]{English}}
        & \texttt{\textcolor{violet!100!black}{retrieve}} & 0.7173 & 0.4024 \\
        & \texttt{\textcolor{red!60!black}{random}} & 0.7265$_{\pm 0.0003}$ & 0.4627$_{\pm 0.0015}$ \\
        & \texttt{\textcolor{blue!60!black}{topic}} & 0.7325$_{\pm 0.0002}$ & 0.5024$_{\pm 0.0005}$ \\
        & \texttt{\textcolor{green!60!black}{similarity}} & 0.7378$_{\pm 0.0006}$ & 0.5159$_{\pm 0.0005}$ \\
        & \texttt{\textcolor{orange!100!black}{re-rank}} & 0.6958 & 0.5503 \\
        & \texttt{\textcolor{cyan!100!black}{llm re-rank}} & \textbf{0.7661} & \textbf{0.5919} \\
        \midrule
        \multirow[c]{6}{*}{\shortstack[l]{Hindi}}
        & \texttt{\textcolor{red!60!black}{random}} & 0.8214$_{\pm 0.0010}$ & 0.4531$_{\pm 0.0010}$ \\
        & \texttt{\textcolor{violet!100!black}{retrieve}} & 0.8005 & 0.4817 \\
        & \texttt{\textcolor{blue!60!black}{topic}} & 0.8328$_{\pm 0.0001}$ & 0.4974$_{\pm 0.0001}$ \\
        & \texttt{\textcolor{green!60!black}{similarity}} & 0.8423$_{\pm 0.0006}$ & 0.5060$_{\pm 0.0001}$ \\
        & \texttt{\textcolor{orange!100!black}{re-rank}} & 0.8036 & 0.5817 \\
        & \texttt{\textcolor{cyan!100!black}{llm re-rank}} & \textbf{0.8507} & \textbf{0.6335} \\
        \bottomrule
    \end{tabular}
    }
    \caption{\textbf{Monolingual} and \textbf{crosslingual} MRR@10 performance across negative sampling strategies (\texttt{random}, \texttt{topic}, \texttt{similarity}) for fine-tuned models on the \textbf{test} set compared to unsupervised strategies (\texttt{retrieve}, cross-encoder \texttt{re-rank}, \texttt{llm re-rank}) for the most represented languages in terms of post--fact-check pairs (i.e., English and Hindi) using the \texttt{multilingual-e5-large} model. Results are ordered by increasing crosslingual score; for the supervised setup, we report averages with standard deviation over three runs using different seeds.}
    \label{tab:english-hindi-cross}
\end{table}

\subsection{Comparing Unsupervised and Supervised Approaches}
\label{sec:results:comparison}

Overall, by comparing the \textit{test} results for the unsupervised and supervised approaches obtained by the best shared model (i.e., \texttt{multilingual-e5-large}, see Table~\ref{tab:supervised_mrr} and Table~\ref{tab:supervised_success_at_10} in Appendix~\ref{sec:appendix:additional}), we observe that in both the multilingual and crosslingual setups the best results in terms of MRR@10 score are obtained by using LLM-based re-ranking (0.8042 and 0.6140, respectively). This is confirmed also by S@10 performance, with 0.8324 and 0.7283 S@10 for the multilingual and crosslingual setups, respectively.
The second best approach is cross-encoder re-ranking, which proves effective in producing better rankings than supervised approaches, especially in the crosslingual setup (0.5442 MRR@10). However, we observe that fine-tuning with negative examples sampled using the \texttt{similarity} strategy leads to better MRR@10 and S@10 performance than cross-encoder re-ranking in the multilingual setup (0.7768 MRR@10 and 0.8228 S@10, respectively).
All other approaches follow these two, with plain unsupervised retrieval showing the worst (or the second worst) results in both scenarios, namely obtaining MRR@10 scores of 0.7475 in the multilingual setup and 0.4323 in the crosslingual setup, and S@10 scores of 0.7971 in the multilingual setup and 0.6324 in the crosslingual setup. 
As regards \texttt{paraphrase-multilingual-mpnet-base-v2}, results confirm that fine-tuning with \texttt{similarity}-sampled negatives is the best strategy in the supervised scenario (0.5537 and 0.3500 MRR@10 and 0.6320 and 0.5157 S@10 in multilingual and crosslingual settings, respectively), but retrieval in an unsupervised fashion provides better performance (0.5690 and 0.3547 MRR@10 and 0.6416 and 0.5188 S@10), ranking after cross-encoder re-ranking (0.6485 and 0.4482 MRR@10 and 0.6796 and 0.5779 S@10). We speculate that this could be due to the small parameter size of this model, which could limit the learning of nuanced patterns from negative and positive pairs.

\begin{figure}
    \resizebox{\linewidth}{!}{%
    
        \begin{tikzpicture}
            \begin{axis}[
                xlabel={$k$ (\# negative examples)},
                ylabel={MRR@10},
                xmin=0.5, xmax=10.5,
                ymin=0.42, ymax=0.50,
                xtick={1,2,3,4,5,10},
                ytick={0.43,0.44,0.45,0.46,0.47,0.48,0.49},
                height=5cm,
                width=9.5cm,
                legend style={at={(0.5,-0.38)}, anchor=north, legend columns=-1},
                ymajorgrids=true,
                grid style=dashed,
            ]

            \addplot[
                color=red!75!white,
                mark=triangle*,
                ]
                coordinates {
                (1,0.4279)(2,0.4312)(3,0.4367)(4,0.4361)(5,0.4353)(10,0.4416)
                };
            \addlegendentry{random}

            \addplot[
                color=green!60!black,
                mark=square*,
                ]
                coordinates {
                (1,0.4583)(2,0.4664)(3,0.4712)(4,0.4747)(5,0.4803)(10,0.4947)
                };
            \addlegendentry{similarity}

            \addplot[
                color=blue!60!black,
                mark=*,
                ]
                coordinates {
                (1,0.4689)(2,0.4725)(3,0.4768)(4,0.4832)(5,0.4863)(10,0.4893)
                };
            \addlegendentry{topic}

            \draw[dashed] (1,0.4323) -- (10,0.4323);
            \node at (8,0.4260) {\emph{without fine-tuning}};
            \end{axis}
        \end{tikzpicture}
    
}%
	\caption{\textbf{Crosslingual} MRR@10 performance across negative sampling strategies and number of negative examples $k$ for the fine-tuned \texttt{multilingual-e5-large} model on the \textbf{test} set. Reported results are averages over three runs using different seeds. The dashed line indicates results when no fine-tuning is conducted.}
	\label{fig:negatives-n-and-strategies-e5-test-mrr}
\end{figure}

Moreover, in both unsupervised and supervised approaches, we observe a notable difference in the contribution of re-ranking and negative sampling, with a much more significant benefit in the crosslingual context. This shared pattern highlights the specificity of the crosslingual context and will be the subject of further investigation.

Overall, our experiments show that fine-tuning embeddings models with negative sampling leads to significant performance improvements, more evident for the crosslingual setup, but requires enough training data for sampling and implies an additional computational effort in the fine-tuning step. Unsupervised PFCR performance, instead, is more dependent on the re-ranking method: while cross-encoder re-ranking underperforms the supervised approach, LLM-based re-ranking yields the best overall performance, but comes at a high computational cost. On the other hand, the unsupervised approach does not require training data and scales well on larger amounts of data.

Additionally, we investigate the monolingual performance for English and Hindi, the two most represented post--fact-check language pairs in the dataset (Table~\ref{tab:english-hindi-cross}). Also in this case, empirical observations confirm the already observed trends, namely that supervised approaches (in particular with \texttt{similarity}-based negative sampling) produce better results in a mono- or multilingual setup, whereas unsupervised, re-ranking based strategies prove more effective in the crosslingual setup.

\section{Conclusion}
\label{sec:conclusion}
We carried out an extensive evaluation of unsupervised and supervised PFCR focusing on multilingual and crosslingual settings. We showed that results in the two settings are remarkably different, with crosslingual retrieval being much more challenging. Unsupervised learning with LLM-based re-ranking yields the best results, even outperforming the best supervised approach. Overall, our study highlights the importance of a thorough evaluation of embedding models and the impact of re-ranking and negative sampling on retrieval performance. We believe that this kind of work could guide the development of PFCR systems tailored to fact-checkers' needs (multilingual \emph{vs} crosslingual) and to the available computational resources.

\section*{Limitations}
Despite the extensive set of experiments and comparisons, this work still has some limitations. A general issue affecting datasets like MultiClaim, which are created by merging different fact-checkers' databases, is the possibility that some pairs of posts and fact-checked claims have not been annotated as positive pairs, even if they should. This is mainly due to the fact that different fact-checking agencies tend to better curate their (monolingual) database, while an extensive effort to annotate positive pairs also crosslingually is often lacking. This may affect the performance of the models and the set of sampled negatives.  

Another possible issue is about the different representation of languages in the dataset. As shown in Table \ref{tab:data_lang_combinations}, English posts are paired with fact-checked claims in almost all languages in our dataset, while some others, such as Dutch or Romanian, have only few crosslingual pairs. Our results and findings could change with a different distribution of languages (and cross-lingual pairs) in the data.

Finally, as detailed in Appendix~\ref{sec:appendix:dataset}, we rely on automatic means to detect languages in posts and fact-checked claims. Although we combine several approaches to increase detection robustness and have also corrected outlier cases (e.g., when Latin was recognized as a language for posts and fact-checks), there might still be some noise in the assigned languages in the data.

\section*{Ethics Statement}

In this paper, we work with the MultiClaim dataset that has been published for research purposes only. We reuse it in line with its terms and conditions; e.g., we do not re-share the subset of data we used, but publish only the IDs together with information about additional metadata (i.e., identified languages), our defined data splits, a list of identified negative examples, and code to load the dataset and run the models and their fine-tuning.

As a part of the paper, we fine-tune text embedding models for the PFCR task. They are intended to be used as an assistance to human fact-checkers or moderators and not to be used in an automated way to fact-check input claims, i.e., to ascertain their veracity based on the retrieved claims.

\section*{Acknowledgments}

This work was partially funded by the European Media and Information Fund (grant number 291191). The sole responsibility for any content supported by the European Media and Information Fund lies with the author(s) and it may not necessarily reflect the positions of the EMIF and the Fund Partners, the Calouste Gulbenkian Foundation and the European University Institute. It was also partially supported by the European Union under the Horizon Europe project AI-CODE, GA No. \href{https://doi.org/10.3030/101135437}{101135437}, the Slovak Research and Development Agency under the project Modermed, GA No. APVV-22-0414, and the PNRR project FAIR – Future AI Research (PE00000013), under the NRRP MUR program funded by NextGeneration EU.

The authors also wish to acknowledge the TAILOR project funded by the European Union under the EU Horizon 2020, GA No.~\href{https://doi.org/10.3030/952215}{952215}, which supported the research mobility that started the collaboration on this paper under the TAILOR Connectivity fund.

\bibliography{anthology,references}

\newpage

\appendix

\section*{Appendix}
\section{Dataset}
\label{sec:appendix:dataset}

\paragraph{Pre-processing} The MultiClaim v2 dataset consists of posts, fact-checked claims, and their pairs. We preprocess the dataset to curate a subset of data for multilingual and crosslingual PFCR. More specifically, we work with posts' (anonymized) texts (\texttt{post\_body} field) and fact-checked claims (\texttt{claim} field), both in their original languages. We thus omit texts extracted with OCR from the images associated with some of the posts because they are too noisy, as well as the titles of the fact-checks, as these are often missing or duplicate the information already present in the claims. We work with pairs, whose \texttt{relationship} is identified either as \texttt{claim\_review} or \texttt{backlink}, since only these relationships were present in the original version of MultiClaim and they represent the ground truth mappings provided by the fact-checkers.

MultiClaim v2 uses Google Translate to identify the languages of posts and fact-checks. However, this being a third party black-box API, we implement our own pipeline to identify languages using open-source tools to have more control and increase robustness of the predictions. Specifically, we use a combination of four language detectors: \emph{i)} \texttt{fastText},\footnote{\url{https://fasttext.cc/docs/en/language-identification.html}} which supports 176 languages~\cite{joulin2016fasttext,joulin2016bag}, \emph{ii)} \texttt{gCLD3} (Compact Language Detector v3),\footnote{\url{https://github.com/google/cld3}} which supports more than 100 languages, \emph{iii)} \texttt{langdetect},\footnote{\url{https://pypi.org/project/langdetect/}} which supports 55 languages, and \emph{iv)} \texttt{polyglot},\footnote{\url{https://polyglot.readthedocs.io/en/latest/}} which supports 196 languages. We combine the outputs of these detectors as follows: we first filter out detected languages that appear only once. Then, we average the normalized detection scores for the remaining ones and filter out those whose average score is $<$0.5. Finally, we take the language with the highest average score as the post/claim detected language.

We also filter out posts whose languages did not appear in at least 200 posts (leading to a cut-off of 180 posts after additional filtering to ensure non-overlapping fact-checks in the data splits; see Section~\ref{sec:dataset}). Yet, we took all fact-checked claims associated with these remaining posts irrespective of their language -- as a result, there are more languages in claims than in posts. We manually checked the languages of claims that appeared $<$10 times (e.g., Esperanto, Latin, Welsh, Corsican). Since these were all misclassifications, we manually corrected the language identified in these cases.

\paragraph{Covered languages} Our subset of the dataset covers 47 languages in total: 30 languages in posts and 46 languages in fact-checked claims. All posts' languages appear in claims except for Urdu, which is represented only in posts. We obtain 283 language combinations in total (see Figure~\ref{tab:data_lang_combinations}). The full list of covered languages is the following:

Afrikaans (\texttt{af}), Arabic (\texttt{ar}), Assamese (\texttt{as}), Azerbaijani (\texttt{az}), Bulgarian (\texttt{bg}), Bengali (\texttt{bn}), Bosnian (\texttt{bs}), Catalan (\texttt{ca}), Czech (\texttt{cs}), Danish (\texttt{da}), German (\texttt{de}), Modern Greek (\texttt{el}), English (\texttt{en}), Spanish (\texttt{es}), Persian (\texttt{fa}), Finnish (\texttt{fi}), French (\texttt{fr}), Hindi (\texttt{hi}), Croatian (\texttt{hr}), Hungarian (\texttt{hu}), Indonesian (\texttt{id}), Italian (\texttt{it}), Kazakh (\texttt{kk}), Korean (\texttt{ko}), Macedonian (\texttt{mk}), Malayalam (\texttt{ml}), Malay (\texttt{ms}), Burmese (\texttt{my}), Nepali (\texttt{ne}), Dutch (\texttt{nl}), Norwegian (\texttt{no}), Punjabi (\texttt{pa}), Polish (\texttt{pl}), Portuguese (\texttt{pt}), Romanian (\texttt{ro}), Russian (\texttt{ru}), Sinhala (\texttt{si}), Slovak (\texttt{sk}), Slovenian (\texttt{sl}), Serbian (\texttt{sr}), Telugu (\texttt{te}), Thai (\texttt{th}), Tagalog (\texttt{tl}), Turkish (\texttt{tr}), Ukranian (\texttt{uk}), Urdu (\texttt{ur}), and 
Chinese (\texttt{zh}).

\section{Unsupervised Approach}
\label{sec:appendix:unsupervised}
In this section, we report the complete, per-model experimental results for the unsupervised setup. Tables~\ref{tab:model_comparison_multilingual} and~\ref{tab:model_comparison_crosslingual} report the results for the baseline retrieval and cross-encoder re-ranking, in both the \textit{test} set and the \textit{full} set of claims. Figures~\ref{fig:s10_retrieve_rerank_performance_test_multi} and~\ref{fig:s10_retrieve_rerank_performance_test_cross} report the same S@10 results in graphical format, also integrating LLM-based re-ranking for the three models involved. 
Figures~\ref{fig:s10_rerank_performance_test_multi} to~\ref{fig:mrr_rerank_performance_full_cross} instead focus on the S@10 and MRR@10 difference in performance (delta) between base retrieval and cross-encoder re-ranking on the \textit{test} and \textit{full} set, respectively. Moreover, Table~\ref{tab:correlation_analysis} reports the full results for the correlation analysis embedding dimension/model dimension \textit{vs.}~model performance.

\begin{table*}[ht!]
\centering
\resizebox{1\linewidth}{!}{
\setlength{\tabcolsep}{3pt} %
\renewcommand{\arraystretch}{1.1} %
\begin{tabular}{l|rrrrrrrrrrrrrrrrrrrrrrrrrrrrrr}
\toprule
\textbf{Post $\rightarrow$} & \texttt{\textbf{ar}} & \texttt{\textbf{bg}} & \texttt{\textbf{bn}} & \texttt{\textbf{cs}} & \texttt{\textbf{de}} & \texttt{\textbf{el}} & \texttt{\textbf{en}} & \texttt{\textbf{es}} & \texttt{\textbf{fr}} & \texttt{\textbf{hi}} & \texttt{\textbf{hr}} & \texttt{\textbf{hu}} & \texttt{\textbf{id}} & \texttt{\textbf{it}} & \texttt{\textbf{ko}} & \texttt{\textbf{mk}} & \texttt{\textbf{ml}} & \texttt{\textbf{my}} & \texttt{\textbf{nl}} & \texttt{\textbf{pl}} & \texttt{\textbf{pt}} & \texttt{\textbf{ro}} & \texttt{\textbf{ru}} & \texttt{\textbf{si}} & \texttt{\textbf{sk}} & \texttt{\textbf{th}} & \texttt{\textbf{tl}} & \texttt{\textbf{tr}} & \texttt{\textbf{ur}} & \texttt{\textbf{zh}} \\
\textbf{Claim $\downarrow$} &  &  &  &  &  &  &  &  &  &  &  &  &  &  &  &  &  &  &  &  &  &  &  &  &  &  &  &  &  &  \\
\bottomrule
\end{tabular}}
\caption{Combinations of the languages in pairs of posts (columns) and fact-checked claims (rows) across all data splits. Languages are represented via their ISO 639-1 two-letter codes. Full language names are in Appendix~\ref{sec:appendix:dataset}.}
\label{tab:data_lang_combinations}
\end{table*}

\section{Supervised Approach}
\label{sec:appendix:supervised}

We report hyper-parameter values in Table~\ref{tab:hp}. The search space was: learning rate: \{1e-9, 5e-9, 1e-8, 5e-8, 1e-7\}, batch size: \{4, 8, 16\}, warm-up steps: \{800, 1,600\}, \# negatives ($k$): \{1, 2, 3, 4, 5, 10\}. Further details are provided in Section~\ref{sec:eval:supervised}.

In Figure~\ref{fig:negatives-k-and-strategies-mpnet}, we present multilingual S@10 results across negative sampling strategies and number of negative examples for \texttt{mpnet} on the \emph{development} set. Moreover, in Figure~\ref{fig:negatives-n-and-strategies-e5-test} we report the crosslingual S@10 performance on the \emph{test} set for \texttt{multilingual-e5-large}, according to negative sampling strategies and number of negatives.

\section{Additional Results}
\label{sec:appendix:additional}

Table~\ref{tab:supervised_success_at_10} shows S@10 \emph{test} scores for all approaches.

Finally, since prior works showed the benefits of translation of posts and fact-checks to English over the use of original data with multilingual models (see e.g., ~\citealp{pikuliak-etal-2023-multilingual}), we also provide results of the unsupervised \texttt{multilingual-e5-large} model with the data translated to English for comparison and completeness despite our focus on multi- and crosslinguality. The achieved score of 0.6996 MRR@10 and 0.7647 S@10 (on the test set in the multilingual setting) show that it is outperformed when using the data with its original languages in contrast to the results by~\citet{pikuliak-etal-2023-multilingual}. This demonstrates the recent progress in the multilingual text embedding models. We hypothesize that the original language can help to retrieve the correct fact-check, especially in the case when both post and fact-check are in the same language. On the other hand, the results on the test set in the crosslingual setting -- 0.6017 MRR@10 (the second best when compared to the MRR@10 results in Table~\ref{tab:supervised_mrr}) and 0.7514 S@10 (the best when compared to the S@10 results in Table~\ref{tab:supervised_success_at_10}) -- show that translation can still help when the original language of the post and the fact-check differ.

\begin{table*}%
\centering
\resizebox{\textwidth}{!}{%
\begin{tabular}{lcc|cc||cc|cc}
\toprule
\multirow{3}{*}{\textbf{Model}} & \multicolumn{4}{c||}{\textbf{Retrieve}} & \multicolumn{4}{c}{\textbf{Re-rank}} \\
& \multicolumn{2}{c|}{\textbf{Test}} & \multicolumn{2}{c||}{\textbf{Full}} & \multicolumn{2}{c|}{\textbf{Test (\textit{top-n = 30})}} & \multicolumn{2}{c}{\textbf{Full (\textit{top-n = 30})}} \\
 & S@10 & MRR@10 & S@10 & MRR@10 & S@10 & MRR@10 & S@10 & MRR@10 \\
\midrule
\texttt{multilingual-MiniLM-L12-v2} & 0.5762 & 0.5062 & 0.4550 & 0.3829 & 0.6950 & 0.6554 & 0.5037 & 0.4561\\
\texttt{paraphrase-multilingual-mpnet-base-v2} & 0.6416 & 0.5690 & 0.5180 & 0.4283 & 0.6796 & 0.6485 & 0.5714 & 0.5068\\
\texttt{e5-mistral-7b-instruct} & 0.7516 & 0.6798 & 0.6257 & 0.5509 & 0.7811 & 0.7223 & 0.6631 & 0.5811\\
\texttt{bge-m3} & 0.7525 & 0.6952 & 0.6586 & 0.5599 & 0.7692 & 0.7031 & 0.6752 & 0.5686\\
\texttt{Linq-Embed-Mistral} & 0.7630 & 0.7199 & 0.6662 & 0.6027 & 0.7731 & 0.7174 & 0.6836 & 0.5952\\
\bottomrule
\end{tabular}
}
\caption{Performance of retrieval and retrieval+re-ranking (\texttt{bge-reranker-v2-m3}) across models in the \textit{test} set and the \textit{full} set of claims in the \textbf{multilingual} setting.}
\label{tab:model_comparison_multilingual}
\end{table*}

\begin{table*}%
\centering

\resizebox{\textwidth}{!}{%
\begin{tabular}{lcc|cc||cc|cc}
\toprule
\multirow{3}{*}{\textbf{Model}} & \multicolumn{4}{c||}{\textbf{Retrieve}} & \multicolumn{4}{c}{\textbf{Re-rank}} \\
& \multicolumn{2}{c|}{\textbf{Test}} & \multicolumn{2}{c||}{\textbf{Full}} & \multicolumn{2}{c|}{\textbf{Test (\textit{top-n = 30})}} & \multicolumn{2}{c}{\textbf{Full (\textit{top-n = 30})}} \\
 & S@10 & MRR@10 & S@10 & MRR@10 & S@10 & MRR@10 & S@10 & MRR@10 \\
\midrule
\texttt{Linq-Embed-Mistral} & 0.3292 & 0.2147 & 0.1865 & 0.1138 & 0.4152 & 0.3410 & 0.2410 & 0.1994 \\
\texttt{multilingual-MiniLM-L12-v2} & 0.3737 & 0.2479 & 0.2225 & 0.1429 & 0.4497 & 0.3512 & 0.2617 & 0.1980 \\
\texttt{text-embedding-3-small} & 0.3899 & 0.2488 & 0.2463 & 0.1434 & 0.4658 & 0.3506 & 0.2855 & 0.1930\\
\texttt{e5-mistral-7b-instruct} & 0.4751 & 0.2837 & 0.2647 & 0.1513 & 0.5840 & 0.4501 & 0.3354 & 0.2608 \\
\texttt{paraphrase-multilingual-mpnet-base-v2} & 0.5188 & 0.3547 & 0.3315 & 0.2048 & 0.5779 & 0.4482 & 0.4137 & 0.3002 \\
\bottomrule
\end{tabular}
}
\caption{Performance of retrieval and retrieval+re-ranking (\texttt{bge-reranker-v2-m3}) across models in the \textit{test} set and the \textit{full} set of claims in the \textbf{crosslingual} setting.}
\label{tab:model_comparison_crosslingual}
\end{table*}

\begin{figure*}%
\centering

%
\caption{S@10 performance of retrieval and re-ranking across models in the \textbf{test} set in the \textbf{multilingual} setting.}
\label{fig:s10_retrieve_rerank_performance_test_multi}
\end{figure*}

\begin{figure*}[!t]
\centering
%
\caption{S@10 performance of retrieval and re-ranking across models in the \textbf{test} set in the \textbf{crosslingual} setting.}
\label{fig:s10_retrieve_rerank_performance_test_cross}
\end{figure*}

\begin{figure}%
    \centering
    \vspace{19pt}
\resizebox{1\columnwidth}{!}{
%
}
    \caption{Difference in \colorbox{blue!30!white}{\textbf{S@10}} \textbf{test} set performance for re-ranking - retrieval in the \textbf{multilingual} setting, with \texttt{bge-reranker-v2-m3}. 6,375 pairs, 5,683 posts, and 6,145 claims. Top-$n=30$.}
    \label{fig:s10_rerank_performance_test_multi}
\end{figure}

\begin{figure}%
    \centering
\resizebox{1\columnwidth}{!}{
%
}
    \caption{Difference in \colorbox{teal}{\textbf{\textcolor{white}{MRR@10}}} \textbf{test} set performance for re-ranking - retrieval in the \textbf{multilingual} setting, with \texttt{bge-reranker-v2-m3}. 6,375 pairs, 5,683 posts, and 6,145 claims. Top-$n=30$.}
    \label{fig:mrr_rerank_performance_test_multi}
\end{figure}

\begin{figure}%
    \centering
    \vspace{18pt}
\resizebox{1\columnwidth}{!}{
%
}
    \caption{Difference in \colorbox{blue!30!white}{\textbf{S@10}} \textbf{full} set performance for re-ranking - retrieval in the \textbf{multilingual} setting, with \texttt{bge-reranker-v2-m3}. 63,913 pairs, 55,421 posts, and 52,911 claims. Top-$n=30$.}
    \label{fig:s10_rerank_performance_full_multi}
\end{figure}

\begin{figure}%
    \centering
    \vspace{12pt}
    \resizebox{1\columnwidth}{!}{
%
}
    \caption{Difference in \colorbox{teal}{\textbf{\textcolor{white}{MRR@10}}} \textbf{full} set performance for re-ranking - retrieval in the \textbf{multilingual} setting, with \texttt{bge-reranker-v2-m3}. 63,913 pairs, 55,421 posts, and 52,911 claims. Top-$n=30$.}    
    \label{fig:mrr_rerank_performance_full_multi}
\end{figure}

\begin{figure}%
    \centering
    \vspace{-12pt}
\resizebox{1\columnwidth}{!}{
%
}
    \caption{Difference in \colorbox{blue!30!white}{\textbf{S@10}} \textbf{test} set performance for re-ranking - retrieval in the \textbf{crosslingual} setting, with \texttt{bge-reranker-v2-m3}. 929 pairs, 850 posts, and 901 claims. Top-$n=30$.}
    \label{fig:s10_rerank_performance_test_cross}
\end{figure}

\begin{figure}%
    \centering
    \vspace{-31pt}
\resizebox{1\columnwidth}{!}{
%
}
    \caption{Difference in \colorbox{teal}{\textbf{\textcolor{white}{MRR@10}}} \textbf{test} set performance for re-ranking - retrieval in the \textbf{crosslingual} setting, with \texttt{bge-reranker-v2-m3}. 929 pairs, 850 posts, and 901 claims. Top-$n=30$.}
    \label{fig:mrr_rerank_performance_test_cross}
\end{figure}

\begin{figure}%
    \centering
    \resizebox{1\columnwidth}{!}{
%
}
    \caption{Difference in \colorbox{blue!30!white}{\textbf{S@10}} \textbf{full} set performance for re-ranking - retrieval in the \textbf{crosslingual} setting, with \texttt{bge-reranker-v2-m3}. 9,066 pairs, 7,975 posts, and 7,869 claims. Top-$n=30$.}
    \label{fig:s10_rerank_performance_full_cross}
\end{figure}

\begin{figure}%
    \centering
    \resizebox{1\columnwidth}{!}{
%
}
    \caption{Difference in \colorbox{teal}{\textbf{\textcolor{white}{MRR@10}}} \textbf{full} set performance for re-ranking - retrieval in the \textbf{crosslingual} setting, with \texttt{bge-reranker-v2-m3}. 9,066 pairs, 7,975 posts, and 7,869 claims. Top-$n=30$.}
    \label{fig:mrr_rerank_performance_full_cross}
\end{figure}

\begin{table}%
    \centering
    \resizebox{1.0\linewidth}{!}{%
    \begin{tabular}{lr}
        \toprule
        \textbf{Hyper-parameter} & \textbf{Value} \\
        \midrule
        Optimizer & AdamW \\
        Epochs & 3 \\
        Batch size & 8 \\
        Learning rate & 1e-8 \\
        Warm-up steps & 1,600 \\
        Label smoothing & 0.1 \\
        Similarity function & cosine \\
        Similarity scale & 20 \\
        Weight decay & 8e-5 \\ 
        Decay factor & 0.38 \\ 
        Cut fraction & 0.3 \\ 
        Clip value & 1 \\
        \midrule
        \# negatives ($k$) & 10 \\
        Sampling strategy & \texttt{similarity}, \texttt{topic}, \texttt{random} \\
		\bottomrule
	\end{tabular}
        }%
	\caption{\label{tab:hp} Final hyper-parameter values used for the supervised models in all our experiments.}
\end{table}

\begin{figure}%
    \resizebox{\linewidth}{!}{%
    
        %
    
}%
	\caption{\textbf{Multilingual} S@10 performance across negative sampling strategies and number of negative examples $k$ for the fine-tuned \texttt{paraphrase-multilingual-mpnet-base-v2} model on the \textbf{development} set. Reported results are averages over three runs using different seeds. The dashed line indicates results when no fine-tuning is conducted.}
	\label{fig:negatives-k-and-strategies-mpnet}
\end{figure}

\begin{figure}[!t]
    \resizebox{\linewidth}{!}{%
    
        %
    
}%
	\caption{\textbf{Crosslingual} S@10 performance across negative sampling strategies and number of negative examples $k$ for the fine-tuned \texttt{multilingual-e5-large} model on the \textbf{test} set. Reported results are averages over three runs using different seeds. The dashed line indicates results when no fine-tuning is conducted.}
	\label{fig:negatives-n-and-strategies-e5-test}
\end{figure}

\newpage

\begin{table}
\centering
\resizebox{1\columnwidth}{!}{
\begin{tabular}{lcc|cc}
    \multicolumn{5}{c}{\textbf{(A) Multilingual - Retrieval}} \\
\toprule
    & \multicolumn{2}{c}{\textbf{test}} & \multicolumn{2}{c}{\textbf{full}} \\
    & \textit{r} & p-value & \textit{r} & p-value \\
\midrule
Emb.~dimension & 0.306 & 0.286 & 0.327 & 0.252 \\
Model size (params) & 0.330 & 0.249 & 0.348 & 0.222 \\
\bottomrule\\[0.75em]
    \multicolumn{5}{c}{\textbf{(B) Multilingual - Re-rank}} \\
\toprule
    & \multicolumn{2}{c}{\textbf{test}} & \multicolumn{2}{c}{\textbf{full}} \\
    & \textit{r} & p-value & \textit{r} & p-value \\
\midrule
Emb.~dimension & 0.374 & 0.188 & 0.376 & 0.185  \\
Model size (params) & 0.372 & 0.191 & 0.370 & 0.193 \\
\bottomrule\\[0.75em]
    \multicolumn{5}{c}{\textbf{(C) Crosslingual - Retrieval}} \\
\toprule
    & \multicolumn{2}{c}{\textbf{test}} & \multicolumn{2}{c}{\textbf{full}} \\
    & \textit{r} & p-value & \textit{r} & p-value \\
\midrule
Emb.~dimension & -0.244 & 0.401 & -0.270 & 0.352  \\
Model size (params) & -0.120 & 0.684 & -0.145 & 0.620 \\
\bottomrule\\[0.75em]
    \multicolumn{5}{c}{\textbf{(D) Crosslingual - Re-rank}} \\
\toprule
    & \multicolumn{2}{c}{\textbf{test}} & \multicolumn{2}{c}{\textbf{full}} \\
    & \textit{r} & p-value & \textit{r} & p-value \\
\midrule
Emb.~dimension & -0.197 & 0.500 & -0.185 & 0.525 \\
Model size (params) & -0.100 & 0.735 & -0.073 & 0.803 \\
\bottomrule
\end{tabular}
}
\caption{\label{tab:correlation_analysis} Correlation analysis between embedding dimension / model size (in terms of number of parameters) and performance (MRR@10) in the unsupervised setting. We report Pearson \textit{r} and the respective p-value. The scores have been computed on the 16 embedding models described in Section \ref{sec:eval:unsupervised}.}
\end{table}

\newpage

\begin{table*}
    \centering
    \small
    
    \begin{tabular}{lllll}
        \toprule
        \textbf{Model} & \textbf{Strategy} & \textbf{Multilingual} & \textbf{Crosslingual} & \textbf{Crosslingual (no eng)}\\
        \midrule
        \multirow[c]{6}{*}{\shortstack[l]{\texttt{multilingual-}\\\texttt{e5-large}}}
        & \texttt{\textcolor{violet!100!black}{retrieve}} & 0.7971 & 0.6324 & 0.6622 \\
        & \texttt{\textcolor{orange!100!black}{re-rank}} & 0.8075 & 0.7069 & \textbf{0.7780} \\
        & \texttt{\textcolor{red!60!black}{random}} & 0.8134$_{\pm 0.0002}$ & 0.6580$_{\pm 0.0012}$ & 0.6853$_{\pm 0.0000}$ \\ 
        & \texttt{\textcolor{blue!60!black}{topic}} & 0.8203$_{\pm 0.0002}$ & 0.7038$_{\pm 0.0000}$ & 0.7224$_{\pm 0.0000}$ \\
        & \texttt{\textcolor{green!60!black}{similarity}} & 0.8228$_{\pm 0.0001}$ & 0.7076$_{\pm 0.0000}$ & 0.7270$_{\pm 0.0000}$ \\ 
        & \texttt{\textcolor{cyan!100!black}{llm re-rank}} & \textbf{0.8324} & \textbf{0.7283} & 0.7409 \\
        
        \midrule
        \multirow[c]{5}{*}{\shortstack[l]{\texttt{paraphrase-}\\\texttt{multilingual-}\\\texttt{mpnet-base-v2}}}
        
        & \texttt{\textcolor{red!60!black}{random}} & 0.5955$_{\pm 0.0002}$ & 0.4758$_{\pm 0.0008}$ & 0.5788$_{\pm 0.0000}$ \\
        & \texttt{\textcolor{blue!60!black}{topic}} & 0.6206$_{\pm 0.0003}$ & 0.5014$_{\pm 0.0012}$ & 0.5819$_{\pm 0.0027}$ \\
        & \texttt{\textcolor{green!60!black}{similarity}} & 0.6320$_{\pm 0.0002}$ & 0.5157$_{\pm 0.0000}$ & 0.5880$_{\pm 0.0000}$ \\
        & \texttt{\textcolor{violet!100!black}{retrieve}} & 0.6416 & 0.5188 & 0.5834 \\
        & \texttt{\textcolor{orange!100!black}{re-rank}} & \textbf{0.6796} & \textbf{0.5779} & \textbf{0.6900} \\
        \bottomrule
        \end{tabular}
    \caption{\label{tab:supervised_success_at_10} \textbf{Multilingual}, \textbf{crosslingual}, and \textbf{crosslingual (no eng)} S@10 performance across negative sampling strategies (\texttt{random}, \texttt{topic}, \texttt{similarity}) for fine-tuned models on the \textbf{test} set compared to unsupervised strategies (\texttt{retrieve}, cross-encoder \texttt{re-rank}, \texttt{llm re-rank}). Results are ordered by increasing multilingual score; for the supervised setup, we report averages with standard deviation over three runs using different seeds.}
\end{table*}

\end{document}